# A Hybrid Evolutionary Approach To Solve University Course Allocation Problem

*Dibyo Fabian Dofadar, Riyo Hayat Khan, Shafqat Hasan, Towshik Anam Taj, Arif Shakil, Mahbub Majumdar*
Department of Computer Science and Engineering
BRAC University
Dhaka, Bangladesh
e-mail: {dibyo.fabian.dofadar, riyo.hayat.khan, shafqat.hasan, towshik.anam.taj}@g.bracu.ac.bd,
{arif.shakil, majumdar}@bracu.ac.bd

*Abstract*—This paper discusses various types of constraints, difficulties and solutions to overcome the challenges regarding university course allocation problem. A hybrid evolutionary algorithm has been defined combining Local Repair Algorithm and Modified Genetic Algorithm to generate the best course assignment. After analyzing the collected dataset, all the necessary constraints were formulated. These constraints manage to cover the aspects needed to be kept in mind while preparing clash free and efficient class schedules for every faculty member. The goal is to generate an optimized solution which will fulfill those constraints while maintaining time efficiency and also reduce the workload of handling this task manually. The proposed algorithm was compared with some base level optimization algorithms to show the better efficiency in terms of accuracy and time.

## I. INTRODUCTION

University departmental tasks are the most crucial parts of a university system as it is interconnected to institutional management, efficiency and effectiveness. It includes inventory management, generating class schedules, assigning Teaching Assistants, handling the process of changing labs of students' due to clashes in their schedule etc. The work shown in this paper focuses on solving the University Course Allocation Problem (UCAP). This is known as an NP-hard problem [1][2] and it involves assignment of faculty members to their respective courses and timeslots, while satisfying a set of constraints. The task of course allocation is quite time consuming when it is done manually. The integration of Artificial Intelligence (AI) has been instrumental in tackling this problem. However, there is no universal solution to this problem since different institutions have different requirements. Researchers have approached this problem by using various optimization algorithms. In this paper, the authors have used a Hybrid Algorithm (HA) which is a combination of Local Repair Algorithm (LRA) and Modified Genetic Algorithm (MGA). Using this algorithm, an optimal solution can be found in the least possible time.

Research in course allocation has been carried out for a long time. Various algorithms such as graph coloring methods, constraint-based methods, population-based methods, metaheuristic methods, variable neighborhood search, hybrid and hyperheuristic approaches have been used by researchers. Population-based methods consist of Genetic Algorithm (GA), Ant-colony Optimization, Memetic Algorithm and metaheuristic methods include Tabu Search (TS) and Simulated Annealing (SA) [1]. In recent years, both population based and local-area based methods have been used to solve this problem. Among them, GA, which is a population based method, has gained popularity because of its high optimization efficiency. Also, SA has been preferred by many researchers since it can avoid getting stuck at the local optimum [3]. GA has some certain advantages over other algorithms. It does not require complex mathematical formulation. Although GA provides globally optimal solutions for complex search spaces [4], it needs more time to execute [5]. Several researchers have reduced this execution time by modifying genetic and heuristic operators and integrating Local Search (LS) techniques [1]. By improving the configuration of several resources, time consumed to reach the optimal solution has been significantly lower compared to conventional GA [6][7][8]. Another way of approaching the UCAP is to use Modified GA and Cooperative GA. Modified GA can produce results at a shorter time and Cooperative GA can reduce the cost value [8][9]. Quality of generated timetable increases during the initial phase of GA according to [10]. The authors proposed to use the generated output of GA as input to another optimizing algorithm to enhance the efficiency. More recent works on GA have proposed modifications to genetic operators such as the crossover operator [11]. Another variation of GA has been introduced by [12] that uses multiple levels of GA computation which the authors have called MDGA (Multi-Depth GA). These levels were based on the depth of the objective function such as shallow, medium and deep. Their goal was to divide the problem to smaller ones to solve the timetabling problem while reducing its time consumption.

Taking the good properties of local and global area based algorithms, [13] have integrated GA with LS algorithms such as SA, TS, Randomized Iterative Local Search (RILS). Their target was to take advantage of the exploration ability of GA and exploitation ability of LS. The LS algorithms help GA to get out from the local optimum. Additionally, fuzzy logic was implemented to check soft constraint violation of the fitness function. Considering the fitness and execution time, hybrid of GA and TS have generated the best optimal solution even when the dataset was large. The hybrid of GA and SA have performed worse than the hybrid of GA and TS, but better than the hybrid of GA and RILS. Recently, [14] has presented a hybrid of

Parallel Genetic Algorithm with LS. The Parallel GA was used to increase the convergence speed and to diversify the population. The solution provided by GA has been improved by minimizing soft constraint violation using LS and the elitism operator. These prevent the GA from getting stuck in the local optimum and lead to better performance.

This paper presents a solution to the UCAP using a Hybrid Algorithm. Section II contains the methodology of this research. Section III discusses the implementation results and analysis of the proposed algorithm. Finally, Section V concludes the paper with the limitations and future scopes of this work.

## II. METHODOLOGY

### A. Data Collection & Pre-processing

The data was collected from The Department of CSE of BRAC University. There are two datasets: Course list of Spring-2020 semester and expression of interest (EOI) responses of every faculty member. Necessary steps were taken for data pre-processing. For example, in the course list dataset, to differentiate from theory course to lab, an extra character 'T' or 'L' was added with the course code. Also, it was found that there can be 36 possible slots in a week. So, the routine for each class was mapped from the value of range 0 to 35 where each value represents a unique day-time slot. Other irrelevant information was scrapped. In this way, the datasets were modified which were later used in the implementation.

### B. Constraints

Before the implementation process, all the constraints were classified into two categories, hard and soft constraints. In total, five hard constraints were introduced. Those are mentioned below:
- No theory section can exist without having a faculty member.
- No lab section can exist without having at least two faculty members.
- No faculty member should have more than one class in each slot.
- No faculty member can take courses exceeding his or her maximum allowed credit limit.
- No faculty members should be given courses outside of their own preferred list, generated from EOI.

Five soft constraints were considered to make the allocation more relevant. For each soft constraint violation, some penalty was added so that the allocation deviates from having a perfect score. The soft constraints are as follows:
- A faculty member should work a maximum of five out of six working days in a week.
- A faculty member should not take more than four slots out of six in a day.
- A faculty member should not take four consecutive slots in a day.
- Several sub-constraints were introduced for different slots in order to minimize the idle time for each faculty member.
- Professors and senior faculty members should not be assigned to any lab classes or any early hour slots.

### C. Evaluation Function

After analyzing the datasets and the constraints, it was decided that it would be much more feasible to use a score-based evaluation approach than using a conventional fitness function. The evaluation function used in the problem was designed in such a way that all the hard constraints must be satisfied and for each soft constraint violation, there should be some penalties based on the significance of that constraint as mentioned earlier. This function has been represented in (1):

$$evaluate(solution) = I(hcv = 0)\frac{\sum_{i=1}^{n_f} max\left(0, \; 1 - \sum_{j=1}^{scv} p_{ij}\right)}{n_f} \qquad (1)$$

Here, $I(hcv = 0)$ is the unit function which keeps track of the hard constraint violation (hcv). In the evaluation process, at any step, for any hard constraint violation count greater than 0, the function will generate 0. Otherwise, it will be set to 1. Then, $n_f$ denotes the total number of faculty members. Initially, for each faculty member, the individual score is set to 1. For every soft constraint violation, the penalty value is subtracted from this individual score. Here, $p_{ij}$ denotes the penalty for the $j^{th}$ soft constraint violation of the $i^{th}$ faculty member. If the summation of these values becomes negative, then it will be set to 0. The final evaluation score of the solution is simply the average of all these individual scores.

### D. Proposed Algorithm

Before the implementation of HA, an initial solution was generated. The optimization process was initiated using that initial solution. This initial solution satisfied all the hard constraints and could be further optimized with the help of the evaluation function used in this algorithm. To generate this solution, for each course index in the course list, a randomized faculty was allocated satisfying all the hard constraints. Additionally, a 'type' indicator was added between a course and a faculty element. Separators were used before and after the type indicator. The structure of each individual of the solution list is shown Fig.1.

First of all, in LRA, this initial solution was assigned to the solution variable and the faculty members were allocated in the courses based on the solution individuals. Then, in a loop, an individual element was selected randomly at first. A suitable faculty member was chosen based on the course index of that element. Then this chosen faculty member index later replaced the current assigned faculty member. This process was done in another loop where the selected element in each iteration would later be modified with a new faculty member. The evaluation function in (1) was used to generate the score for each assignment. The maximum score was updated if the score found in the current iteration appeared to be better than the previous maximum score and the solution was also updated. After the inner loop ended, based on the solution the suitable faculty was assigned in the selected individual. This whole process was done for all the course-faculty elements, until the stopping criterion was met. After the termination, the algorithm returned the intermediate solution. The function LRA gives an overview of what has been described so far.

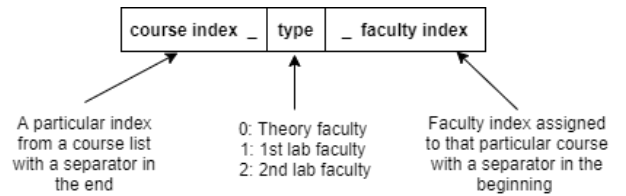

Figure 1. Structure of an Individual in Solution List

```
function LRA(initial_solution):
    while current_iteration < total_iteration do
        selected ← a random individual from
                   current_solution
        for f in all possible faculties of selected course do
            selected ← replace previous selected with f
            if evaluate(current_solution) >
            evaluate(previous_solution) then
                solution ← current_solution
            end if
        end for
    end while
    return solution
end function
```

Next in MGA, the algorithm took the intermediate solution as a parameter. A loop was started with the condition of stopping the algorithm when the maximum numbers of generations were reached. In the loop, a random course was selected at the beginning and then based on the selected course, the crossover operation occurred. In this algorithm, the crossover operation is referred as 'mini-batch crossover' where a small size of random course-faculty individuals was taken from the current solution. In this operation, the faculty index was interchanged between the selected course-faculty elements. After several successful crossover operations, if the current score was not updating after a certain tolerance limit, the boolean variable activate_mutation was set to true. Unlike conventional approaches, the mutation operation in this algorithm does not start automatically. Rather it depends on the mutation tolerance limit. That is why this operation is referred in this algorithm as 'conditional mutation'. In this operation, after crossover, some random faculty elements of some individuals were modified based on the course index. Then the modified elements were inserted into the current solution and the score was generated using the evaluation function afterwards. The maximum score was updated if the new score was found to be better than the previous maximum score and the solution was also updated. Once the maximum iteration was reached, the best solution for this algorithm was returned. The function MGA gives an overview of what has been described so far.

```
function MGA(initial_solution):
    while current_generation < max_generation do
        mini_batch ← select an even number of random
                     individuals from the solution
        mini_batch_crossover(mini_batch_size)
        if activate_mutation is True then
            elem ← random element from modified_selected
            conditional_mutation(elem)
        end if
        if evaluate(current_solution) >
        evaluate(previous_solution) then
            solution ← current_solution
        end if
    end while
    return solution
end function
```

The proposed hybrid algorithm is the combination of these above two algorithms. This hybrid algorithm is proved to be faster than normal evolutionary approaches. In the algorithm, the initial solution was passed as the parameter in the function LRA. This process is faster but due to the absence of evolutionary steps like crossover and mutation, this algorithm does not provide a better score and gets stuck in the local maxima. For that reason, a switching_tolerance factor was introduced where it was set as true whenever the score for the current solution was not updating for a certain number of iterations. After that, the function MGA started with the intermediate solution as the argument. In MGA, the score updated rapidly when the crossover operation was run. This score was further optimized with the help of the conditional mutation operation. Thus, the final optimized solution was returned after the execution of the HA. This process is mentioned in the function HA.

```
function HA(initial_solution)
    LRA(initial_solution)
    if switching_tolerance exceeds then
        MGA(intermediate_solution)
    end if
    return solution
end function
```

## III. RESULT AND ANALYSIS

### A. Parameter Settings

TABLE I. PARAMETER SETTINGS

| Algorithm | Parameter Name | Parameter Value |
|---|---|---|
| LRA | Total Iteration | 5000 |
| LRA | Switching Tolerance | 1000 |
| MGA | Total Generation | 500000 |
| MGA | Mini-Batch Crossover Size | 2 |
| MGA | Mutation Tolerance | 20000 |

### B. Analysis of a Test Case

The algorithm was run several times. The final score ranged from 0.91 to 0.93 and the total time taken varied from 1670 to 2290 seconds for these particular datasets.

Among these test cases, the visualization of the best one is shown in Fig 2. This figure shows the score vs time graph for that test case.

From the figure, it is seen that, when LRA runs, the score increases rapidly. At a certain point, it gets stuck in the local maxima, hence the score remains unchanged. When the switching_tolerance value exceeds (intermediate phase), then the algorithm switches to MGA. Here, at first only because of the crossover operation, the score increases once again. This score can be further modified when activate_mutation value becomes true and mutation operation starts.

### C. Comparison with Base Level Algorithms

To determine the efficiency of this algorithm more precisely, a comparison with some other common algorithms such as conventional population based algorithms (Genetic Algorithm, Memetic Algorithm) and metaheuristic approaches (Stochastic Hill Climbing, Simulated Annealing, Tabu Search) was required. For that reason, those commonly used algorithms were implemented using the same datasets. Table II shows the comparison of the accuracy of the implemented algorithm with these algorithms with the same initial solution and evaluation function. The accuracy is nothing but the percentage of the final score. Here, all the algorithms were run several times too and the best result is mentioned in the table.

If we visualize these outputs, we can clearly see the differences among these algorithms in terms of both time and accuracy. Fig. 3 shows that HA outperforms all other algorithms and gives the best accuracy in a minimum amount of time.

TABLE II. COMPARISON WITH BASE LEVEL ALGORITHMS

| Algorithm Name | Total Time Taken (sec) | Accuracy (%) |
|---|---|---|
| Proposed Hybrid Algorithm | 1671 | 92.99 |
| Genetic Algorithm | 4579 | 89.89 |
| Stochastic Hill Climbing | 2831 | 84.32 |
| Simulated Annealing | 2905 | 85.15 |
| Tabu Search | 2866 | 85.98 |
| Memetic Algorithm | 4731 | 90.98 |

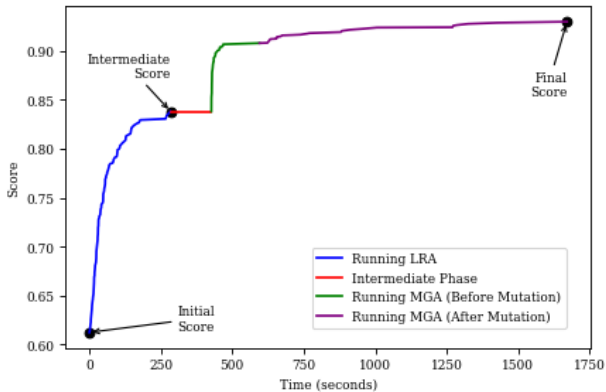

Figure 2. Score vs Time Graph of a Test Case for the Proposed Algorithm

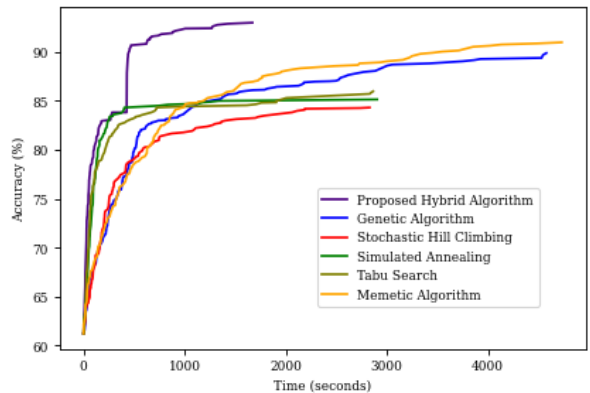

Figure 3. Comparison Between Different Algorithms

## IV. Conclusion

The integration of HA in this work has surpassed the performance of other base level algorithms in terms of the final result. The hybridization of algorithms should be explored more frequently as it can enhance the effectiveness of the problem solving techniques compared to when the algorithms are used individually. This research can be improved further and more areas of the problem can be covered. Other optimization algorithms like Particle Swarm Optimization and Ant-Colony Optimization can also be used for solving this problem. Different combinations of optimization algorithms can be utilized to prepare robust hybrid algorithms, which is both cost efficient and less time consuming. Also, the prepared preferred course list of the dataset does not include any type of priority. A priority list of courses can be introduced in further modifications. Another solution to this problem can be in terms of setting a maximum course limit for each faculty member for a particular course. This limit will determine how many sections of this course one faculty member can take in a semester. Strategies similar to this idea can also be used to solve other departmental tasks such as Teaching Assistant appointment and Lab Change procedure of the students. The authors hope that this work will open new doors of opportunities for the researchers who are working with similar topics and through further research, they will be able to take this idea to a more advanced level where the management will be simpler and easier.

## Acknowledgment

The authors would like to thank the Almighty for keeping us healthy in these tough times. The authors would also like to show their gratitude towards the Department of Computer Science & Engineering of BRAC University for providing us with the necessary data to complete this research.